\documentclass[10pt,twocolumn,letterpaper]{article}

\usepackage[pagenumbers]{cvpr} 

\usepackage{graphicx}
\usepackage{amsmath}
\usepackage{amssymb}
\usepackage{booktabs}

\usepackage[pagebackref,breaklinks,colorlinks]{hyperref}

\usepackage[capitalize]{cleveref}
\usepackage{multirow}
\usepackage{booktabs}
\crefname{section}{Sec.}{Secs.}
\Crefname{section}{Section}{Sections}
\Crefname{table}{Table}{Tables}
\crefname{table}{Tab.}{Tabs.}

\def\cvprPaperID{94} 
\def\confName{CVPR}
\def\confYear{2022}

\begin{document}

\title{Self-Calibrated Efficient Transformer for Lightweight Super-Resolution}

\author{Wenbin Zou$^{1, *}$, Tian Ye$^{2, *}$, Weixin Zheng$^{3, }$\thanks{Equal contribution}, Yunchen Zhang$^4$, Liang Chen$^{1, }$\thanks{Corresponding author}, Yi Wu$^1$\\
Fujian Provincial Key Laboratory of Photonics Technology, Fujian Normal University, Fuzhou, China.$^1$\\
School of Ocean Information Engineering, Jimei University, Xiamen, China.$^2$\\
College of Physics and Information Engineering, Fuzhou University, Fuzhou, China.$^3$\\
China Design Group Co., Ltd., Nanjing, China.$^4$\\
{\tt\small alexzou14@foxmail.com, 201921114031@jmu.edu.cn, 	visinzheng@163.com,}\\
{\tt\small cydiachen@cydiachen.tech, cl\_0827@126.com, wuyi@fjnu.edu.cn}
}
\maketitle

\begin{abstract}
   Recently, deep learning has been successfully applied to the single-image super-resolution (SISR) with remarkable performance. However, most existing methods focus on building a more complex network with a large number of layers, which can entail heavy computational costs and memory storage. To address this problem, we present a lightweight Self-Calibrated Efficient Transformer (SCET) network to solve this problem. The architecture of SCET mainly consists of the self-calibrated module and efficient transformer block, where the self-calibrated module adopts the pixel attention mechanism to extract image features effectively. To further exploit the contextual information from features, we employ an efficient transformer to help the network obtain similar features over long distances and thus recover sufficient texture details. We provide comprehensive results on different settings of the overall network. Our proposed method achieves more remarkable performance than baseline methods. The source code and pre-trained models are available at \url{https://github.com/AlexZou14/SCET}.
\end{abstract}

\section{Introduction}
\label{sec:intro}

\begin{figure}
	\centering
	\includegraphics[width=8.2cm]{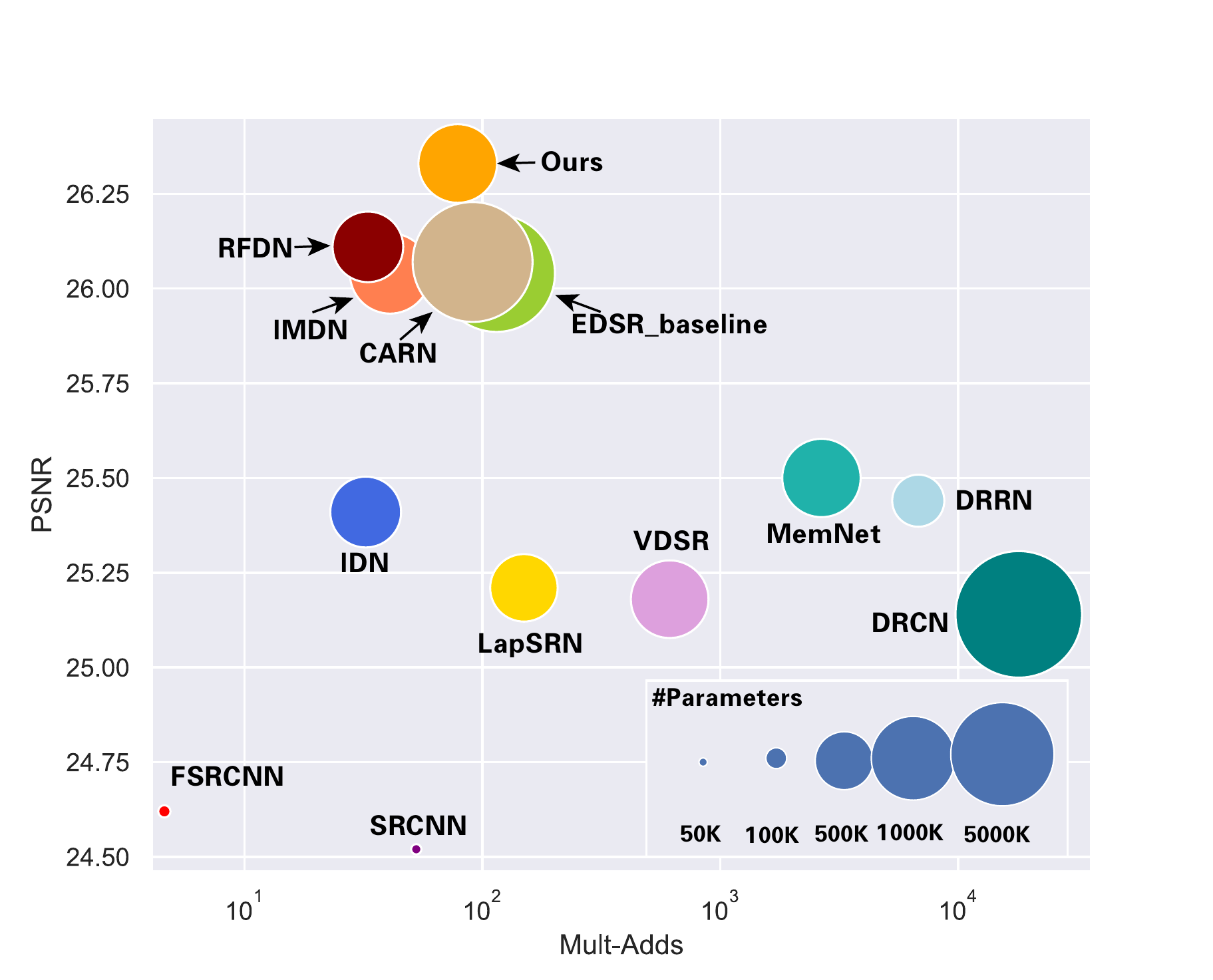}\\
	\caption{Trade-off between performance vs: number of operations and parameters on Urban100 $\times$4 dataset. Multi-adds are calculated on 720p HR image. The results show the superiority of our model among existing methods.}
	\label{ComparsionMultiAdd}
\end{figure}

Single image super-resolution (SISR) \cite{SISR} aims to recover a high-resolution (HR) image from its low-resolution (LR) observation, which is a challenging ill-posed problem because many latent HR images can be downsampled to an identical LR image. To address this significant problem, many image super-resolution (SR) methods \cite{SRCNN, VDSR, EDSR} based on deep convolution architecture have been proposed and shown impressive performance. Thanks to the powerful representation capabilities of the deep convolution neural networks, numerous previous approaches can learn the complex non-linear mapping from paired LR-HR images. 

Dong \textit{et al.} \cite{SRCNN} firstly propose the super-resolution convolutional neural network (SRCNN) that outperforms the previous work. On this basis, various SR algorithms \cite{FSRCNN, VDSR, DRCN, DRRN} have been proposed with superior performances, and those methods have a large margin compared with traditional methods. It is widely known that deeper networks based on residual learning \cite{ResNet} generally achieve better performances. Based on this cognition, deeper networks with larger frameworks, e.g. enhanced deep super-resolution network (EDSR) \cite{EDSR} and residual channel attention network (RCAN) \cite{RCAN}, have been proposed and achieved excellent performance. However, previous CNN-based SR networks have a large number of parameters, resulting in the limitation of the application of SR technology in edge devices. 

A straightforward solution to this problem is to design lightweight and efficient networks via reducing the amount of the parameters, \textit{e.g.}, building shallow networks with a single path \cite{FSRCNN, LapSRN}, recursive operation \cite{DRCN,DRRN}, information distillation mechanism \cite{IDN, IMDN}, and neural architecture search (NAS) \cite{NAS1, NAS2}. However, most of these methods focus on local contextual information and do not consider global similar textures, leading to problems such as artifacts in the recovered image. The limited receptive field of convolution operation is difficult to capture globally similar features, resulting in a poor trade-off between performance and complexity.

The image restoration methods based on the transformer architecture have made remarkable progress recently. Yet, there are few studies on the lightweight SR transformer network, which attracts us to explore the following exciting topic:
~\\ ~\\
~\textit{How to design a \textbf{lightweight} transformer to \textbf{effectively} perform single image super-resolution?} 
~\\ ~\\
Previous distillation-based solutions achieve impressive SISR performance. However, the above solutions have redundant parameters as the channel-splitting design of extract features progressively in a single basic block. Furthermore, they still have scope for improvement in performance as the spatial and channel modeling ability is relatively weak. 

According to the above analysis, the core idea of our approach is how to make lightweight networks with both spatial modeling and channel modeling capabilities. Due to the complexity limitations, it is obviously more efficient to model dependencies in the channel and spatial dimensions respectively. Thus, we propose two complementary components, the SC module and the efficient transformer module to endow the network with powerful modeling capabilities in the spatial dimension and channel dimension respectively.

~\textit{Self-Calibrated Module.} We propose the SC module as the efficient extractor to explore the valuable spatial features from low-resolution input. With the help of the spatial attention mechanism, it adaptively pays more attention to the detailed textures. Therefore, the SC module provides strong spatial clues for the following transformer module. 

~\textit{Efficient Transformer Module.} We construct a linear-complexity transformer module to perform channel-wise self-attention mechanism, which efficiently models the dependence in the channel dimension from input features. The combination of two proposed modules provides complementary clues in the channel and spatial dimensions for the HR image reconstruction. 

Based on above components, we propose a lightweight \textbf{S}elf-\textbf{C}alibrated \textbf{E}fficient \textbf{T}ransformer (\textbf{SCET}) network to solve the SISR problem efficiently. For instance, our method achieves higher performance than the state-of-the-art lightweight SR method A$^2$F-M \cite{A2F} with 0.53 dB PSNR gain on the $\times$4 Manga109 \cite{manga109} dataset, the number of parameters in SCET only 68.3\% of A$^2$F-M. The SCET method is a competing entry in NTIRE 2022 Efficient Super-Resolution challenge \cite{li2022ntire}.





The key contributions of this work are as follows:
\begin{itemize}
    \item We introduce the efficient transformer design to the lightweight SISR task, effectively exploiting to the property that the transformer module can capture long-range dependencies, avoiding the problem of wrong textures generated by current lightweight SR methods.
    \item We design the SC module as the high-performance extractor. Compared with the information distillation mechanism in the IMDB block \cite{IMDN}, the SC module employs a more efficient feature propagation strategy, achieving better performance with fewer parameters and less computational effort.
    \item As shown in Figure \ref{ComparsionMultiAdd}, our SCET occupies fewer parameters and takes fewer Multi-Adds, while significantly improving the performance of SISR networks at low resource consumption.
\end{itemize}

\section{Related Work}

\subsection{Deep SR models}
In recent years, deep CNN is employed in various low-level vision tasks, such as image denoising \cite{denosie}, deblurring \cite{deblur}, and so on. Dong \textit{et al.} \cite{SRCNN} make a big step forward by proposing a three-layer fully convolutional network SRCNN. On this basis, Kim \textit{et al.} design deeper network VDSR \cite{VDSR} and DRCN \cite{DRCN} via residual learning. Subsequently, Tai \textit{et al.} \cite{DRRN} later develop a deep recursive residual networks (DRRN) by introducing recursive blocks and then propose a persistent memory network (MemNet) \cite{MemNet} by utilizing memory block. However, the above methods use the bicubic interpolation to preprocess the LR image, which inevitably losses some details and bring large computation. To solve this problem, Dong \textit{et al.} \cite{FSRCNN} propose FSRCNN by adopting a deconvolution layer to upsample images at the end of the network to decrease computations. Then, Shi \textit{et al.} \cite{ESPCNN} introduce an efficient sub-pixel convolutional layer instead of deconvolution. On this basis, Lim \textit{et al.} \cite{EDSR} propose a deeper and wider network EDSR by stacking residual blocks (eliminating batch normalization layers). The significant performance gain indicates the fact that the depth and width of the network occupy important places in image SR. Furthermore, some other networks, e.g. non-local neural network (NLRN) \cite{NLRN}, RCAN \cite{RCAN}, and second-order attention network (SAN) \cite{SAN}, improve the performances by modeling the correlation of features in space or channel dimensions. Yet, these networks sacrifice the portability of the network, leading to the highly cost in memory storage and computational complexity.

\subsection{Lightweight SR models}

During these years, many lightweight networks have been working on SR problem. They can be approximately divided into three classes: the architectural design-based methods \cite{CARN, IDN, LapSRN}, the knowledge distillation-based methods \cite{SRdistill}, and the NAS-based methods \cite{NAS1, NAS2}. The first class mainly focuses on the recursive operation and channel splitting. Deeply-recursive convolutional network (DRCN) \cite{DRCN} and deep recursive residual network (DRRN) \cite{DRRN} are proposed to share parameters via introducing the recursive layers. However, the reduction of computational operation and the amount of parameters are still unsatisfying. Ahn \textit{et al.} design a cascading residual network (CARN) \cite{CARN}, that accomplishes a cascading mechanism based on residual learning. Lightweight feature fusion network (LFFN) \cite{LFFN} uses multi-path channel learning to incorporate multi-scale features. NAS \cite{NAS}, which is an emerging approach to automatically design efficient networks, is introduced to the SR task \cite{NAS1, NAS2}. However, the performances of NAS-based methods are limited by the search space and strategies. IMDN \cite{IMDN} extracts hierarchical features step by step through splitting operations and further improves the efficiency of the model. On this basis, RFDN \cite{RFDN} has further improved the information multi-distillation block in IMDN and won the first place at the Efficient Super-Resolution Challenge in AIM 2020 \cite{AIM2020}. Inspired by SCNet \cite{SCNet}, Zhao \textit{et al.} \cite{PAN} employ a self-calibrated convolution with pixel attention block, which further reduces the network parameters and improves the network operation speed. Therefore, we employ the self-calibrated convolution scheme in our SCET network for efficient SR.

\subsection{Vision Transformer}

The breakthroughs from Transformer in the NLP area lead to sigificant interest in the computer vision community. It has been successfully applied in image recognition \cite{Transrecog1, Transrecog2, li2021localvit}, object detection \cite{Transobj1, Transobj2} and segmentation\cite{Seg1,Seg2}. Currently, most Vision Transformer split the image into a sequence of patches and then flatten them into vectors to learn their interrelationships through self-attention. Therefore, the Vision Transformers possesses the strong capability to learn long-term dependencies between image pixel. Owing to its powerful learning capabilities, Transformer is introduced to low-level vision tasks \cite{IPT, SRT,valanarasu2021transweather,liang2021swinir} and obtained excellent performance recently. However, the self-attention mechanism in the Transformer introduces a huge amount of computation and GPU resource consumption, which is not friendly to lightweight networks. Therefore, building efficient Vision Transformer has become a hot research topic in recent years.

\begin{figure*}
\vspace{-0.3cm}
\setlength{\abovecaptionskip}{0.1cm} 
\setlength{\belowcaptionskip}{-0.5cm}
	\centering
	\includegraphics[width=14cm]{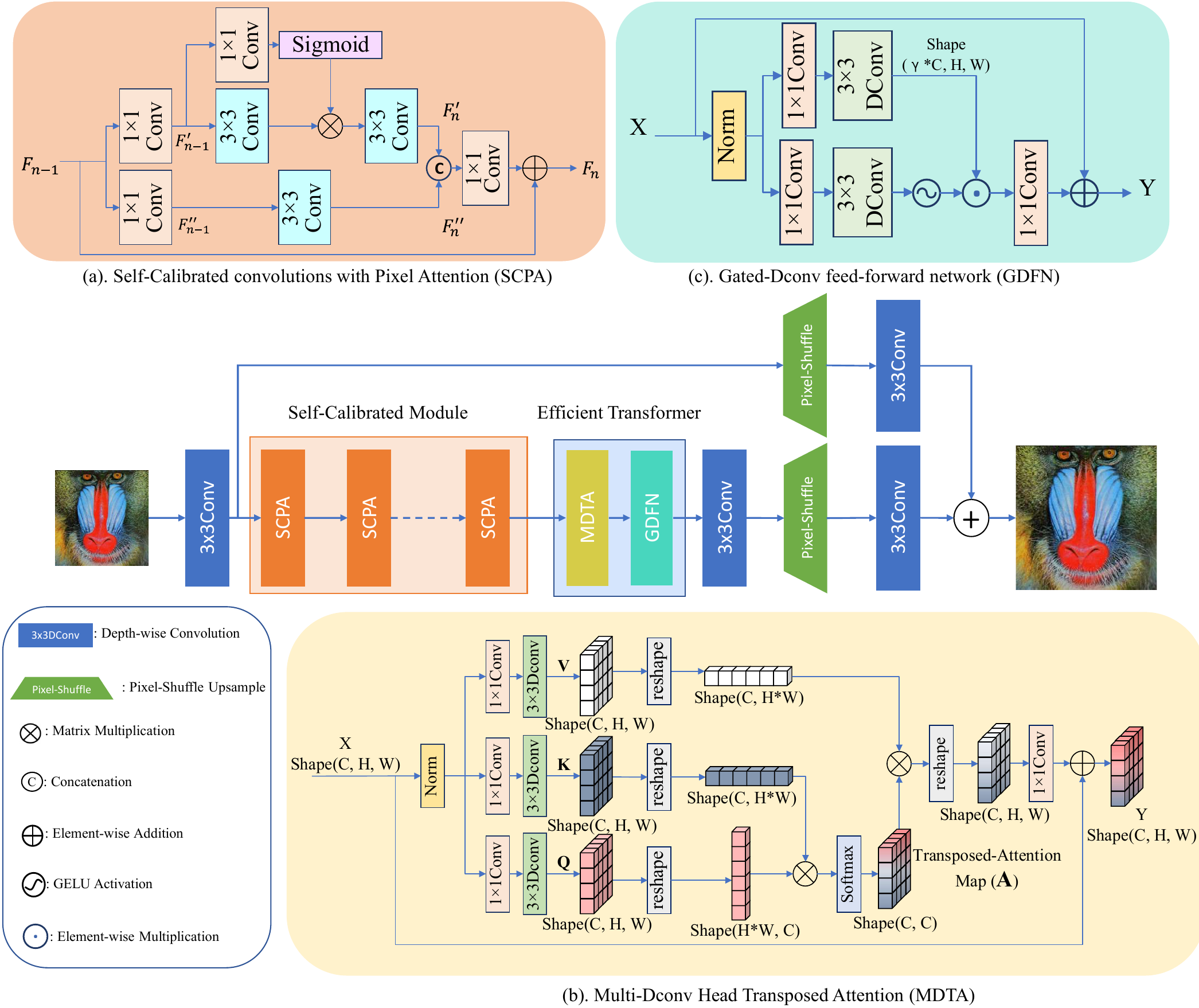}\\
	\caption{The architecture of self-calibrated efficient transformer (SCET) network. Here, the core modules of network are: (a) Self-calibrated convolution with pixel attention (SCPA), (b) Multi-Dconv head transposed attention (MDTA), and (c) Gated-Dconv feed-forward network (GDFN). }
	\label{network}
\end{figure*}

\section{Self-Calibrated Efficient Transformer}

In this section, we present the overall architecture of the proposed Self-Calibrated Efficient Transformer (SCET) firstly. Then, we introduce the lightweight self-calibrated (SC) module, which consists of several stacked self-calibrated convolutions with pixel attention (SCPA) blocks to efficiently extract texture information from images. Finally, we describe the efficient transformer module.

\subsection{Overview of Network Framwork}
Considering that complex network structure blocks may bring a large number of parameters and complexity, we choose a simple network structure, as shown in Figure \ref{network}. Our SCET mainly consists of two parts: SC module and efficient transformer module. Specifically, the SC module is used to efficiently extract image texture features and the Efficient Tranformer module is used to recover similar textures across long ranges.

Given an input low-resolution image $I_{LR} \in \mathbb{R}^{H\times W\times 3}$, SCET first applies a convolution to obtain shallow feature $F_0 \in \mathbb{R}^{H\times W\times C}$, where $H \times W$ denotes the spatial dimension and $C$ is the number of channels. It can be formulated as:
\begin{equation}
    F_0 = H_{conv}(I_{LR}),
\end{equation}
where $H_{conv}$ denotes $3\times 3$ convolution operation. Next, inspired by PAN \cite{PAN}, we employed an SC module composed of SCPA blocks to efficiently extract the deep texture feature. It can be expressed as:
\begin{equation}
    F_{SC}=H_{SC}(F_0),
\end{equation}
where $H_{SC}$ denotes SC module, $F_{SC}$ denotes the output of SC module. To obtain global similarity information, we use the efficient transformer module to further recover similar textures across long distances. Inspired by Restormer \cite{restormer} that the amount of computation can reduce from $\mathcal{O}(W^2H^2)$ to $\mathcal{O}(C^2)$ by applying self-attention to compute cross-covariance across channels, we employ the multi-Dconv head transposed attention (MDTA) to generate an attention map encoding the global context implicitly. Besides, we adopt a gated-Dconv feed-forward network (GDFN) to focus on the fine texture details complimentary. It can be written as:

\begin{equation}
    F_{out} = H_{ET}(F_{SC}) = H_{GDFN}(H_{MDTA}(F_{SC})),
\end{equation}
where $H_{ET}$, $H_{MDTA}$, and $H_{GDFN}$ denote the efficient transformer, MDTA and GDFN, respectively. $F_{out}$ denote the output of efficient transformer. Finally, we utilize the pixel-shuffle to upsample the features to the HR size. In addition, we added a global residual path to make full use of the shallow feature information. It can be expressed as:
\begin{equation}
    I_{SR} = H_{up}^1(F_{out})+H_{up}^2(F_0)=H_{SCET}(I_{LR}),
\end{equation}
where $H_{up}^1$ and $H_{up}^2$ denote the upsampling operation of the backbone network and the upsampling operation of the global residual path. $H_{SCET}$ denotes the proposed SCET network. $I_{SR}$ denotes the final restored image.

\subsection{Self-Calibrated Module}

Most CNN-based lightweight SR networks extract hierarchical features step-by-step to reduce parameters and computational effort, making the insufficient use of low-frequency information resulting in poor image recovery. We employ the SC module constructed from SCPA for feature extraction and recovery. Instead of the step-by-step approach, the SC module allows the network to purposefully recover missing textures through pixel attention. As depicted in Figure \ref{network}, our SC module consists of several SCPA blocks. It can be expressed as:
\begin{equation}
    F_{out} = H_{SCPA}^n(H_{SCPA}^{n-1}(\cdots H_{SCPA}^0(F_{in})\cdots)),
\end{equation}
where $H_{SCPA}^n$ denotes the function of the $n$-th SCPA blocks. $F_{in}$ and $F_{out}$ denote the input and output of the SC module, respectively. Next, we describe specifically the SCPA block in the SC module, as shown in Figure \ref{network} (a). We define $F_{n-1}$ and $F_n$ as the input and output of the $n$-th SCPA blocks, respectively. The SCPA block consists of two branches, one for the computation of pixel attention information and the other for the recovery of spatial domain information directly. Specifically, the SCPA block first uses pixel convolution of the two branches to reduce the half number of channels. It can be written as:
\begin{align}
    F_{n-1}'=H_{pconv}^1(F_{n-1}),\\
    F_{n-1}''=H_{pconv}^2(F_{n-1}),
\end{align}
where $H_{pconv}^1$ and $H_{pconv}^2$ denote the pixel convolution of upper and lower branch, respectively. $F_{n-1}'$ and $F_{n-1}''$ only have half of the channel number of $F_{n-1}$. Then, the upper branch computes the attention information by a pixel attention, and the lower channel branch through a 3$\times$3 convolution to recover the spatial domain information. It can be expressed as:
\begin{align}
    F_{PA} &= H_{conv}(F_{n-1}')\odot \sigma(H_{pconv}(F_{n-1}')),\\
    F_n' &= H_{conv}^1(F_{PA}),\\
    F_n''&=H_{conv}^2(F_{n-1}''),
\end{align}
where $\sigma$ and $\odot$ denote the function of sigmoid and element-wise multiplication, respectively. $F_{PA}$ denotes the pixel attention map. Finally, the output features of the two branches are concatenated together, and then the attention information and spatial domain information are fused together by a pixel convolution to recover the missing texture information in a targeted manner. It can be expressed as:
\begin{equation}
    F_n = H_{pconv}(concat(F_n', F_n'')) + F_n,
\end{equation}
where $concat$ denotes the operation of concatenation. In order to accelerate training, local residual path is used to produce the final output feature $F_n$.

\subsection{Efficient Transformer}
To further improve the performance of our network, we use the efficient transformer module to obtain global contextual information, allowing the network to recover more high frequency texture details. Our efficient transformer consists of MDTA and GDFN. Next, we introduce each module in the efficient transformer in detail.

The major computational overhead in the Transformer lies in the self-attention layer and tends to grow quadratically with the input size. To alleviate this problem, we employ MDTA to compute the cross-covariance over the channel dimensions, as shown in Figure \ref{network} (b). Specifically, we use pixel convolution and depth-wise convolution in three branches to generate query (\textbf{Q}), key (\textbf{K}) and value (\textbf{V}) from the input features $\textbf{X}\in \mathbb{R}^{H\times W\times C}$. It can be expressed as:
\begin{align}
    \textbf{Q} = H_{dconv}^1(H_{pconv}^1(LN(\textbf{X}))),\\
    \textbf{K} = H_{dconv}^2(H_{pconv}^2(LN(\textbf{X}))),\\
    \textbf{V} = H_{dconv}^3(H_{pconv}^3(LN(\textbf{X}))),
\end{align}
where $H_{dconv}$, $H_{pconv}$ and $LN$ denote depth-wise convolution, pixel convolution, and the layer normalization, respectively. Then, we apply the reshape operation to obtain $\hat{\textbf{Q}}\in \mathbb{R}^{HW\times C} $, $\hat{\textbf{K}}\in \mathbb{R}^{C\times HW}$ and $\hat{\textbf{V}}\in \mathbb{R}^{HW\times C}$. Next, their dot-product interaction generates a transposed-attention map \textbf{A} of size $\mathbb{R}^{C\times C}$. It can be defined as:
\begin{align}
    \textbf{A} &= \textbf{V}\cdot Softmax(\textbf{K}\cdot \textbf{Q}/\alpha), \\
    \textbf{Y} &= H_{pconv}(\textbf{A}) + \textbf{X},
\end{align}
where $Softmax$ denotes the function of softmax to generate probability map. $\alpha$ is a learnable scaling parameters to control the magnitude of the dot product of $\textbf{K}$ and $\textbf{Q}$. Unlike the existing Transformer which calculates self-attention on the spatial domain, MDTA can effectively reduce the amount of computation. 

To further recover the accurate structural information, we also adopt the gated-Dconv feed-forward Network. Instead of the feed-forward network in the existing Transformer, GDFN has more operational operations to help the network focus on recovering high frequency details using contextual information, as shown in Figure \ref{network} (c). Given the input feature $\textbf{X} \in \mathbb{R}^{H\times W\times C}$, GDFN can be formulated as:
\begin{align}
    \textbf{X}_{G}^1 &= \phi(H_{dconv}(H_{pconv}(LN(\textbf{X})))),\\
    \textbf{X}_{G}^2 &= H_{dconv}(H_{pconv}(LN(\textbf{X}))),\\
    \textbf{Y}_G &= \textbf{X}_{G}^1 \odot \textbf{X}_{G}^2,\\
    \textbf{Y} &= H_{pconv}(\textbf{Y}_{G}),
\end{align}
where $LN$ and $\phi$ denote layer normalization and the function of GELU. GDFN controls the information flow through the respective hierarchical levels in our method, thereby allowing each level to focus on the fine details complimentary to the other levels.

Overall, our efficient transformer effectively helps the network to obtain global contextual information to recover high frequency texture details.

\subsection{Loss Function}
Our SCET is optimized with mean absolute error (MAE, also known as L1) loss function for a fair comparison. Given a training set $\{I_{LR}^i, I_{HR}^i\}$, that contains $\mathcal{N}$ LR inputs and their HR counterparts. The goal of training SCET is to minimize the $L_1$ loss function:
\begin{equation}
    L(\Theta)=\dfrac{1}{N}\sum_{i=1}^N ||H_{SCET}(I_{LR}^i)-I_{HR}^i||_1,
\end{equation}
where $\Theta$ denotes the parameter set of SCET and $||\cdot||_1$ is $L_1$ norm. The loss function is optimized by using stochastic gradient descent (SGD) algorithm. More training details of our method are presented in Section \ref{sec:experiments}.

\begin{table*}[!]
\vspace{-0.3cm}
\setlength{\abovecaptionskip}{0cm} 
\setlength{\belowcaptionskip}{-0.8cm}
    \caption{Average PSNR/SSIM for scale factor $\times 2$, $\times 3$ and $\times 4$ on datasets Set5, Set14, B100, Urban100, and Manga109. Best and second best results are \textcolor{red}{red} and \textcolor{blue}{blue}}
    \centering
    \resizebox{13cm}{!}{
    \begin{tabular}{c|c|c|c|c|c|c|c}
    \hline
        \multirow{2}{*}{Method} & \multirow{2}{*}{Scale} & \multirow{2}{*}{Params} & Set5 & Set14 & B100 & Urban100 & Manga109  \\ 
         &  & & PSNR/SSIM & PSNR/SSIM & PSNR/SSIM & PSNR/SSIM & PSNR/SSIM  \\ \hline
        Bicubic & \multirow{12}{*}{$\times$2} & - & 33.66/0.9299     & 30.24/0.8688 & 29.56/0.8431 & 26.88/0.8403 & 30.80/0.9339   \\ 
        SRCNN \cite{SRCNN} &  & 8K & 36.66/0.9542 & 32.45/0.9067 & 31.36/0.8879 & 29.50/0.8946 & 35.60/0.9663  \\ 
        VDSR \cite{VDSR} &  & 666K  & 37.53/0.9587     & 33.03/0.9124 & 31.90/0.8960 & 30.76/0.9140 & 37.22/0.9750  \\ 
        DRRN \cite{DRRN} &  & 298K  & 37.74/0.9591 & 33.23/0.9136 & 32.05/0.8973 & 31.23/0.9188 & 37.88/0.9749  \\
        DRCN \cite{DRCN} &  & 1,774K  & 37.63/0.9588     & 33.04/0.9118 & 31.85/0.8942 & 30.75/0.9133 & 37.55/0.9732  \\ 
        IDN \cite{IDN} &  & 553K  & 37.83/0.9600     & 33.30/0.9148 & 32.08/0.8985 & 31.27/0.9196 & 38.01/0.9749  \\ 
        CARN \cite{CARN} & & 1,592K & 37.76/0.9590     & 33.52/0.9166 & 32.09/0.8978 & 31.92/0.9256 & 38.36/0.9765  \\ 
        IMDN \cite{IMDN} &  & 694K  & 38.00/0.9605     & 33.63/0.9177 & 32.19/0.8996 & 32.17/0.9283 & 38.88/0.9774  \\ 
        PAN \cite{PAN} &  & 261K  & 38.00/0.9605     & 33.59/0.9181 & 32.18/0.8997 & 32.01/0.9273 & 38.70/0.9773  \\ 
        RFDN \cite{RFDN} &  & 534K & 38.05/0.9606    & 33.68/0.9184 & 32.16/0.8994 & 32.12/0.9278 & 38.88/0.9773  \\ 
        A$^2$F-M \cite{A2F} & & 999K & \textcolor{blue}{38.04}/\textcolor{blue}{0.9607}     & \textcolor{blue}{33.67}/\textcolor{blue}{0.9184} & \textcolor{blue}{32.18}/\textcolor{blue}{0.8996} & \textcolor{blue}{32.27}/\textcolor{blue}{0.9294} & \textcolor{blue}{38.87}/\textcolor{blue}{0.9774}  \\ 
        SCET (Ours) &  & 683K & \textcolor{red}{38.06}/\textcolor{red}{0.9615}     & \textcolor{red}{33.78}/\textcolor{red}{0.9198} & \textcolor{red}{32.24}/\textcolor{red}{0.9006} & \textcolor{red}{32.38}/\textcolor{red}{0.9299} & \textcolor{red}{39.86}/\textcolor{red}{0.9821} \\
            \hline
        Bicubic & \multirow{12}{*}{$\times$3} & - & 30.39/0.8682         & 27.55/0.7742 & 27.21/0.7385 & 24.46/0.7349 & 26.95/0.8556  \\ 
        SRCNN \cite{SRCNN} &  & 8K &  32.75/0.9090 & 29.30/0.8215 & 28.41/0.7863 & 26.24/0.7989 & 30.48/0.9117  \\ 
        VDSR \cite{VDSR} &  & 666K & 33.66/0.9213          & 29.77/0.8314 & 28.82/0.7976 & 27.14/0.8279 & 32.01/0.9340  \\ 
        DRCN \cite{DRCN} &  & 1,774K & 33.82/0.9226         & 29.76/0.8311 & 28.80/0.7963 & 27.15/0.8276 & 32.24/0.9343  \\ 
        DRRN \cite{DRRN}&  & 298K  & 34.03/0.9244 & 29.96/0.8349 & 28.95/0.8004 & 27.53/0.8378 & 32.71/0.9379  \\
        IDN \cite{IDN} &  & 553K & 34.11/0.9253     & 29.99/0.8354 & 28.95/0.8013 & 27.42/0.8359 & 32.71/0.9381  \\ 
        CARN \cite{CARN} &  & 1,592K & 34.29/0.9255         & 30.29/0.8407 & 29.06/0.8034 & 28.06/0.8493 & 33.50/0.9440  \\ 
        IMDN \cite{IMDN} &  & 703K & 34.36/0.9270         & 30.32/0.8417 & 29.09/0.8046 & 28.17/0.8519 & 33.61/0.9445  \\ 
        PAN \cite{PAN} &  & 261K & 34.40/0.9271         & 30.36/0.8423 & 29.11/0.8050 & 28.11/0.8511 & 33.61/0.9448  \\ 
        RFDN \cite{RFDN} &  & 541K & 34.41/0.9273        & 30.34/0.8420 & 29.09/0.8050 & 28.21/0.8525 & 33.67/0.9449  \\ 
        A$^2$F-M \cite{A2F} &  & 1003K & \textcolor{blue}{34.50}/\textcolor{blue}{0.9278}         & \textcolor{blue}{30.39}/\textcolor{blue}{0.8427} & \textcolor{blue}{29.11}/\textcolor{blue}{0.8054} & \textcolor{blue}{28.28}/\textcolor{blue}{0.8546} & \textcolor{blue}{33.66}/\textcolor{blue}{0.9453}  \\ 
        SCET (Ours) &  & 683K  & \textcolor{red}{34.53}/\textcolor{red}{0.9278}         & \textcolor{red}{30.43}/\textcolor{red}{0.8441} & \textcolor{red}{29.17}/\textcolor{red}{0.8075} & \textcolor{red}{28.38}/\textcolor{red}{0.8559} & \textcolor{red}{34.29}/\textcolor{red}{0.9503}  \\ 
         \hline
        Bicubic & \multirow{12}{*}{$\times$4} & -  & 28.42/0.8104     & 26.00/0.7027 & 25.96/0.6675 & 23.14/0.6577 & 24.89/0.7866  \\ 
        SRCNN \cite{SRCNN} &  & 8K & 30.48/0.8626     & 27.50/0.7513 & 26.90/0.7101 & 24.52/0.7221 & 27.58/0.8555  \\ 
        VDSR \cite{VDSR} &  & 666K & 31.35/0.8838              & 28.01/0.7674 & 27.29/0.7251 & 25.18/0.7524 & 28.83/0.8870  \\ 
        DRCN \cite{DRCN} &  & 1,774K & 31.53/0.8854             & 28.02/0.7670 & 27.23/0.7233 & 25.14/0.7510 & 28.93/0.8854  \\
        DRRN \cite{DRRN} &  & 298K & 31.68/0.8888     & 28.21/0.7720 & 27.38/0.7284 & 25.44/0.7638 & 29.45/0.8946  \\
        IDN \cite{IDN} &  & 553K & 31.82/0.8903         & 28.25/0.7730 & 27.41/0.7297 & 25.41/0.7632 & 29.41/0.8942  \\ 
        CARN \cite{CARN} &  & 1,592K & 32.13/0.8937             & 28.60/0.7806 & 27.58/0.7349 & 26.07/0.7837 & 30.47/0.9084  \\ 
        IMDN \cite{IMDN} &  & 715K & 32.21/0.8948             & 28.58/0.7811 & 27.56/0.7353 & 26.04/0.7838 & 30.45/0.9075  \\ 
        PAN \cite{PAN} &  & 272K & 32.13/0.8948              & 28.61/0.7822 & 27.59/0.7363 & 26.11/0.7854& 30.51/0.9095 \\ 
        RFDN \cite{RFDN} &  & 550K & 32.24/0.8952            & 28.61/0.7819 & 27.57/0.7360 & 26.11/0.7858 & \textcolor{blue}{30.58}/0.9089  \\ 
        A$^2$F-M \cite{A2F} &  & 1010K & \textcolor{red}{32.28}/\textcolor{blue}{0.8955}& \textcolor{blue}{28.62}/\textcolor{blue}{0.7828} & \textcolor{blue}{27.58}/\textcolor{blue}{0.7364} & \textcolor{blue}{26.17}/\textcolor{blue}{0.7892} & {30.57}/\textcolor{blue}{0.9100}  \\
        SCET (Ours) &  & 683K & \textcolor{blue}{32.27}/\textcolor{red}{0.8963} & \textcolor{red}{28.72}/\textcolor{red}{0.7847} & \textcolor{red}{27.67}/\textcolor{red}{0.7390} & \textcolor{red}{26.33}/\textcolor{red}{0.7915} & \textcolor{red}{31.10}/\textcolor{red}{0.9155}  \\ 
         \hline
    \end{tabular}
    }\label{PSNRandSSIM}
\end{table*}

\section{Experiments}
\label{sec:experiments}

\begin{figure*}
\vspace{-0.6cm}
\setlength{\abovecaptionskip}{-0.35cm} 
\setlength{\belowcaptionskip}{-0.8cm}
	\centering
	\includegraphics[scale=0.7,trim={1cm 5cm 0.5cm 0.5cm},clip]{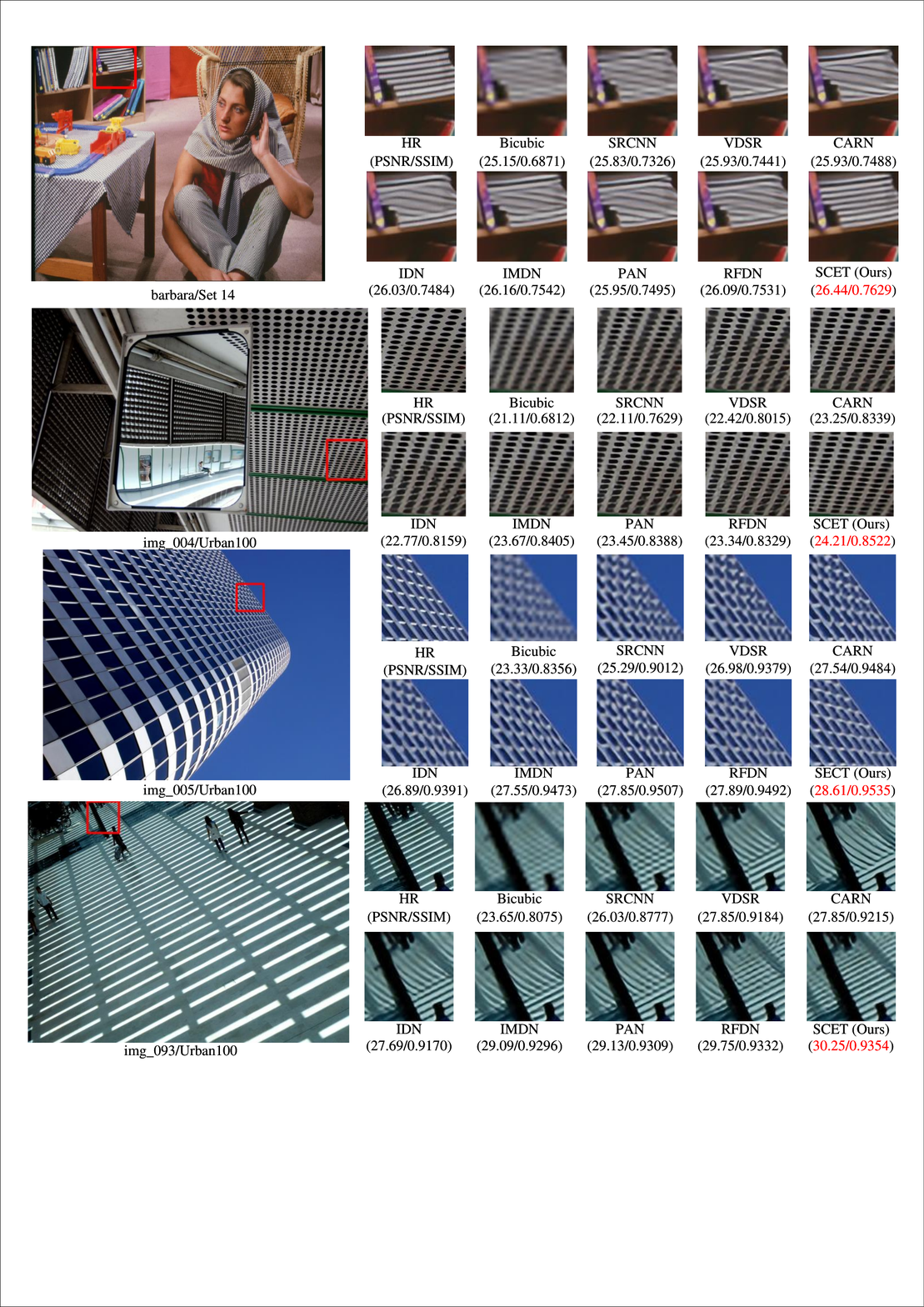}\\
	\caption{Qualitative comparison with the leading algorithms: SRCNN \cite{SRCNN}, VDSR \cite{VDSR}, CARN \cite{CARN}, IDN \cite{IDN}, IMDN \cite{IMDN}, PAN \cite{PAN}, and RFDN \cite{RFDN} on $\times$4 task. From the figure, we can see that our method can generate finer details of the image and achieve outstanding performance.}
	\label{imgcp}
\end{figure*}

\subsection{Settings}
In this subsection, we clarify the experimental setting about datasets, degradation models, evaluation metrics, and training settings.

\textbf{Dataset.} Following the previous methods \cite{IDN, IMDN, PAN, RFDN, A2F}, we conduct the training process on a widely used dataset, DIV2K \cite{DIV2K} and Flickr2K \cite{Flickr2K}, which contains 3450 LR-HR RGB image pairs. We augment the training data with random horizontal flips and rotations. For testing, we use five standard benchmark datasets: Set5 \cite{Set5}, Set14 \cite{Set14}, B100 \cite{B100}, Urban100 \cite{Urban100}, and Manga109 \cite{manga109}.

\textbf{Degradation models.} We downscale HR images with the scaling factors (×2, ×3, and ×4) using Bicubic degradation models \cite{degrade1, degrade2}.

\textbf{Evaluation metrics.} The SR images are evaluated with PSNR and SSIM \cite{PSNR} on Y channel of transformed YCbCr space. Besides, we use Multi-Adds (the size of a query image is $1280 \times 720$) and model parameters to evaluate the computational complexity of a model.

\textbf{Training Settings.} We give the implementation details of the proposed SCET. The numbers of the SCPA blocks and feature channels in the self-calibrated module are flexible and configurable, which set 16 and 64, respectively. During training, We train our model SCET on the crop training dataset with LR and HR, the ground turth patch size is random crop into $416\times416$. We use the Adam \cite{Adam} optimizer with the $2\times 10^{-4}$ learning rate to training 1,000,000 iteration and decay the learning rate with the cosine strategy. Weight decay is $10^{-4}$ for all the training periodic. We implement our model on the PyTorch platform. Training the SCET roughly takes two days with one RTX2080Ti GPU for the whole training.

\subsection{Comparisons with State-of-the-art Methods}
\textbf{Results with Bicubic degaradation.} It is widely used to simulate LR images with Bicubic degradation in image SR settings. To verify the effectiveness of our SCET, we compare SCET with 10 SOTA image SR methods: SRCNN \cite{SRCNN}, VDSR \cite{VDSR}, DRCN \cite{DRCN}, DRRN \cite{DRRN}, IDN \cite{IDN}, CARN \cite{CARN}, IMDN \cite{IMDN}, PAN \cite{PAN}, RFDN \cite{RFDN}, and A$^2$F-M \cite{A2F}. All the quantitative results for various scaling factors are reported in Table \ref{PSNRandSSIM}. Compared with other methods, our SCET, with fewer parameters and computation complexity, performs the best results on five datasets with various scaling factors.

\textbf{Visual Results of Recent Methods.} To further illustrate the superiority of SCET, we also show the visual results of various methods (Bicubic upsampling, SRCNN \cite{SRCNN}, VDSR \cite{VDSR}, CARN \cite{CARN}, IDN \cite{IDN}, IMDN \cite{IMDN}, PAN \cite{PAN}, RFDN \cite{RFDN}, and our SCET) in Figure \ref{imgcp}. We can see that most baseline models cannot reconstruct the lattices accurately and thus suffer from serious aliasing. In contrast, our SCET obtains sharper results and recovers more highfrequency details. Take the image img\_093/Urban100 for example, most compared methods output heavy aliasing. The early developed methods, i.e., Bicubic upsampling, SRCNN \cite{SRCNN}, VDSR \cite{VDSR} and CARN \cite{CARN} lose most of the structure due to the limited network depth and abundant inefficient features. More recent methods, such as IDN \cite{IDN}, IMDN \cite{IMDN}, PAN \cite{PAN}, and RFDN \cite{RFDN}, can recover the main outlines but fail to recover shaper details. Compared with that, our SCET can restore more details and sharper edges and gain higher visual quality. That should be attributed to more efficient feature extraction and the ability to access global information.

\textbf{Model Complexity.} To further prove the ascendency of SCET in terms of complexity, we compare performance in the matter of parameters and computational complexity. As shown in Figure \ref{ComparsionMultiAdd}, SCET with limited operations and performance, achieves better performance than other large models. This shows that SCET has a good balance between model complexity and performance.

\subsection{Ablation Study}
In this subsection, we design a series of ablation experiments to analyze the effectiveness of each of the modules we propose. We use the DIV2K validation dataset for evaluation and performed 1,000,000 iterations of training on an input image patch of size 32 × 32.

\textbf{Model Design Policy.}
We explore the impact of different depths and widths on network performance, as shown in Table \ref{depthandwide}. The depth represents the number of SPCA blocks and the width represents the numberof channels in our intermediate features. As can be seen from the experimental results, the width affects network performance and parameters more than the depth. Our model works best at d = 16 and w = 64. Therefore, our final model is set to d = 16 and w = 64.

\textbf{Comparison of different backbone schemes.} 
To illustrate the effectiveness of the SCPA as a backbone, we used the residual block, residual channel attention block (RCAB), information multi-distillation block (IMDB) and residual feature distillation block (RFDB) to replace the original SCPA blocks for the ablation experiments. 

\begin{table}
	\caption{Model Policy with deep and wide on network performance. The `d' denotes the number of SCPA blocks. The `w' denotes the number of feature channels.}
	\centering
	\begin{tabular}{l|c|c|c|c}
		\hline
		Model       & Params & Multi-Adds  &      PSNR      &      SSIM       \\[1pt] \hline\hline
		d = 8, w = 32&  98k   &  11.46G   &     28.32      &     0.7741      \\[1pt] \hline
		d = 8, w = 64 & 388k  &  44.85G  &   28.64     & 0.7894 \\[1pt] \hline
		d = 16, w = 32 & 172k  & 19.9G  &  28.58      &     0.7869      \\[1pt] \hline
		d = 16, w = 64 &  683k  & 78.72G &  \textbf{28.72}      &     \textbf{0.8158}      \\[1pt] \hline
	\end{tabular}
	\label{depthandwide}
\end{table}

\begin{table}
	\caption{Ablation studies of different backbone. We report the PSNR (dB) values on DIV2K validation datasets ($\times$4).}
	\centering
	\begin{tabular}{l|cccc}
		\hline
		Backbone& Params & Multi-Adds  &      PSNR      &      SSIM       \\[1pt] \hline
		ResBlock& 1274k &  146.87G  &    28.29        &     0.7965       \\[1pt] 
		RCAB &  1284k   &  146.87G  &  28.32  &  0.7984  \\[1pt] 
		IMDB &  920k   &  106.05G &  28.49     &  0.8033    \\[1pt] 
		RFDB &  1336k  &  145.9G  &  28.57  & 0.8042        \\[1pt] 
		SCPA &  683k  &  78.72G  &  28.72  &  0.8158  \\[1pt] \hline
	\end{tabular}
	\label{backbone}
\end{table}

\begin{table}
	\caption{Ablation studies of different transformer. We report the PSNR (dB) values on DIV2K validation datasets ($\times$4).}
	\centering
	\resizebox{8.2cm}{!}{
	\begin{tabular}{l|c|c|c|c}
		\hline
		Transformer       & Component & Params  &      Multi-Adds      &      PSNR       \\[1pt] \hline
		Baseline&  SCPA blocks  &  629K   &     72.59G       &  28.54  \\[1pt] \hline
		\multirow{2}{*}{Self-Attention} &  MTA+FN  &   1002K &  129.65G  &  28.62  \\[1pt] 
		                                &  MDTA+FN &  929K &  107.09G   &  28.69  \\[1pt] \hline
		\multirow{2}{*}{\begin{tabular}[c]{@{}l@{}}Feed-forward \\Network\end{tabular}} &  MDTA+Resblock  & 721K & 83.14G  &  28.59   \\[1pt] 
		                                & MDTA+RCAB & 722K  &  84.21G  &    28.61     \\[1pt] \hline
		Overall                         & MDTA+GDFN &  683K  &     78.72G   &    28.72       \\[1pt] \hline
	\end{tabular}
	}
	\label{ablation}
\end{table}

In Table \ref{backbone}, we give the comparison in terms of parameters, Multi-Adds, and the performance in PSNR. Note that all results are the mean values of PSNR calculated by 100 images on DIV2K validation dataset. Mult-Adds is computed by assuming that the resolution of HR image is 720p. It is observed that SCPA could achieve the best performance with the fewest parameters and Multi-Adds. SCPA can reduce parameters and calculations by nearly half in comparison to RFDB, obtaining a performance improvement of 0.15dB. This indicates that SCPA is more effective than traditional basic modules which employ a step-by-step approach to extract hierarchical features.

\textbf{Comparison of different Transformer schemes.}
To illustrate the effectiveness of MDTA and GDFN in efficient transformer, we compare the effects of different approaches to self-attention and different feed-forward networks on the model. Note that our baseline model is set up as a residual network of cascading multiple SCPA blocks. 

As shown in Table \ref{ablation}, it demonstrates that the MDTA provides favorable gain of 0.18 dB over the baseline. The MDTA can reduce the amount of computation by $20\%$ compared to traditional self-attention. Moreover, it is shown that deep convolution can effectively improve the robustness of the efficient transformer. For feedback networks, the gating mechanism in GDFN that controls the information flowing can effectively help the network to obtain better performance. Compared to other feedforward network designs, the GDFN can improve performance by about 0.1 dB.
\section{Conclusion}
In this paper, we propose a lightweight SCET network for efficient super-resolution. In particular, we design a new Efficient Transfomer framework, which effectively combines the efficient pixel attention mechanism with the transformer to achieves excellent results with few parameters. Additionally, numerous experiments have shown that the proposed method achieves a commendable balance between visual quality and parameters amount, which are the vital factors that affect practical use of SISR.

\section*{Acknowledgments} 
This work was supported in part by the National Nature Science Foundation of China under Grant No. 61901117, U1805262, 61971165, in part by the Natural Science Foundation of Fujian Province under Grant No. 2019J05060, 2019J01271, in part by the Special Fund for Marine Economic Development of Fujian Province under Grant No. ZHHY-2020-3, in part by the research program of Fujian Province under Grant No. 2018H6007, the Special Funds of the Central Government Guiding Local Science and Technology Development under Grant No. 2017L3009, and the National Key Research and Development Program of China under Grant No. 2016YFB1001001.

{\small
\bibliographystyle{ieee_fullname}
\bibliography{ref}

\begin{thebibliography}{10}\itemsep=-1pt

\bibitem{denosie}
Chunwei~Tian A, Yong Xu~A B, and Wangmeng~Zuo C.
\newblock Image denoising using deep cnn with batch renormalization.
\newblock {\em Neural Networks}, 121:461--473, 2020.

\bibitem{CARN}
Namhyuk Ahn, Byungkon Kang, and Kyung-Ah Sohn.
\newblock Fast, accurate, and lightweight super-resolution with cascading
  residual network.
\newblock 03 2018.

\bibitem{Set5}
Marco Bevilacqua, Aline Roumy, Christine Guillemot, and Marie~Line
  Alberi-Morel.
\newblock Low-complexity single-image super-resolution based on nonnegative
  neighbor embedding.
\newblock 2012.

\bibitem{Transobj1}
Nicolas Carion, Francisco Massa, Gabriel Synnaeve, Nicolas Usunier, Alexander
  Kirillov, and Sergey Zagoruyko.
\newblock End-to-end object detection with transformers.
\newblock In {\em European conference on computer vision}, pages 213--229.
  Springer, 2020.

\bibitem{IPT}
Hanting Chen, Yunhe Wang, Tianyu Guo, Chang Xu, Yiping Deng, Zhenhua Liu, Siwei
  Ma, Chunjing Xu, Chao Xu, and Wen Gao.
\newblock Pre-trained image processing transformer.
\newblock In {\em Proceedings of the IEEE/CVF Conference on Computer Vision and
  Pattern Recognition}, pages 12299--12310, 2021.

\bibitem{NAS1}
Xiangxiang Chu, Bo Zhang, Hailong Ma, Ruijun Xu, and Qingyuan Li.
\newblock Fast, accurate and lightweight super-resolution with neural
  architecture search.
\newblock In {\em 2020 25th International Conference on Pattern Recognition
  (ICPR)}, pages 59--64. IEEE, 2021.

\bibitem{NAS2}
Xiangxiang Chu, Bo Zhang, and Ruijun Xu.
\newblock Multi-objective reinforced evolution in mobile neural architecture
  search.
\newblock In {\em European Conference on Computer Vision}, pages 99--113.
  Springer, 2020.

\bibitem{LFFN}
Xiangxiang Chu, Bo Zhang, and Ruijun Xu.
\newblock Multi-objective reinforced evolution in mobile neural architecture
  search.
\newblock In {\em European Conference on Computer Vision}, pages 99--113.
  Springer, 2020.

\bibitem{SAN}
T. {Dai}, J. {Cai}, Y. {Zhang}, S. {Xia}, and L. {Zhang}.
\newblock Second-order attention network for single image super-resolution.
\newblock In {\em 2019 IEEE/CVF Conference on Computer Vision and Pattern
  Recognition (CVPR)}, pages 11057--11066, 2019.

\bibitem{Transobj2}
Xiyang Dai, Yinpeng Chen, Jianwei Yang, Pengchuan Zhang, Lu Yuan, and Lei
  Zhang.
\newblock Dynamic detr: End-to-end object detection with dynamic attention.
\newblock In {\em Proceedings of the IEEE/CVF International Conference on
  Computer Vision}, pages 2988--2997, 2021.

\bibitem{SRCNN}
C. {Dong}, C.~C. {Loy}, K. {He}, and X. {Tang}.
\newblock Image super-resolution using deep convolutional networks.
\newblock {\em IEEE Transactions on Pattern Analysis and Machine Intelligence},
  38(2):295--307, 2016.

\bibitem{FSRCNN}
Chao Dong, Chen~Change Loy, and Xiaoou Tang.
\newblock Accelerating the super-resolution convolutional neural network.
\newblock 08 2016.

\bibitem{Transrecog1}
Alexey Dosovitskiy, Lucas Beyer, Alexander Kolesnikov, Dirk Weissenborn,
  Xiaohua Zhai, Thomas Unterthiner, Mostafa Dehghani, Matthias Minderer, Georg
  Heigold, Sylvain Gelly, et~al.
\newblock An image is worth 16x16 words: Transformers for image recognition at
  scale.
\newblock {\em arXiv preprint arXiv:2010.11929}, 2020.

\bibitem{SISR}
William~T Freeman, Egon~C Pasztor, and Owen~T Carmichael.
\newblock Learning low-level vision.
\newblock {\em International journal of computer vision}, 40(1):25--47, 2000.

\bibitem{SRdistill}
Qinquan Gao, Yan Zhao, Gen Li, and Tong Tong.
\newblock Image super-resolution using knowledge distillation.
\newblock In {\em Asian Conference on Computer Vision}, pages 527--541.
  Springer, 2018.

\bibitem{ResNet}
Kaiming He, Xiangyu Zhang, Shaoqing Ren, and Jian Sun.
\newblock Identity mappings in deep residual networks.
\newblock In {\em European conference on computer vision}, pages 630--645.
  Springer, 2016.

\bibitem{Urban100}
J. {Huang}, A. {Singh}, and N. {Ahuja}.
\newblock Single image super-resolution from transformed self-exemplars.
\newblock In {\em 2015 IEEE Conference on Computer Vision and Pattern
  Recognition (CVPR)}, pages 5197--5206, 2015.

\bibitem{IMDN}
Zheng Hui, Xinbo Gao, Yunchu Yang, and Xiumei Wang.
\newblock Lightweight image super-resolution with information
  multi-distillation network.
\newblock In {\em Proceedings of the 27th acm international conference on
  multimedia}, pages 2024--2032, 2019.

\bibitem{IDN}
Zheng Hui, Xiumei Wang, and Xinbo Gao.
\newblock Fast and accurate single image super-resolution via information
  distillation network.
\newblock In {\em Proceedings of the IEEE conference on computer vision and
  pattern recognition}, pages 723--731, 2018.

\bibitem{VDSR}
J. {Kim}, J.~K. {Lee}, and K.~M. {Lee}.
\newblock Accurate image super-resolution using very deep convolutional
  networks.
\newblock In {\em 2016 IEEE Conference on Computer Vision and Pattern
  Recognition (CVPR)}, pages 1646--1654, 2016.

\bibitem{DRCN}
J. {Kim}, J.~K. {Lee}, and K.~M. {Lee}.
\newblock Deeply-recursive convolutional network for image super-resolution.
\newblock In {\em 2016 IEEE Conference on Computer Vision and Pattern
  Recognition (CVPR)}, pages 1637--1645, 2016.

\bibitem{Adam}
Diederik Kingma and Jimmy Ba.
\newblock Adam: A method for stochastic optimization.
\newblock {\em Computer ence}, 2014.

\bibitem{LapSRN}
W. {Lai}, J. {Huang}, N. {Ahuja}, and M. {Yang}.
\newblock Deep laplacian pyramid networks for fast and accurate
  super-resolution.
\newblock In {\em 2017 IEEE Conference on Computer Vision and Pattern
  Recognition (CVPR)}, pages 5835--5843, 2017.

\bibitem{li2021localvit}
Yawei Li, Kai Zhang, Jiezhang Cao, Radu Timofte, and Luc Van~Gool.
\newblock Localvit: Bringing locality to vision transformers.
\newblock {\em arXiv preprint arXiv:2104.05707}, 2021.

\bibitem{li2022ntire}
Yawei Li, Kai Zhang, Luc~Van Gool, Radu Timofte, et~al.
\newblock Ntire 2022 challenge on efficient super-resolution: Methods and
  results.
\newblock In {\em IEEE Conference on Computer Vision and Pattern Recognition
  Workshops}, 2022.

\bibitem{liang2021swinir}
Jingyun Liang, Jiezhang Cao, Guolei Sun, Kai Zhang, Luc Van~Gool, and Radu
  Timofte.
\newblock Swinir: Image restoration using swin transformer.
\newblock In {\em Proceedings of the IEEE/CVF International Conference on
  Computer Vision}, pages 1833--1844, 2021.

\bibitem{EDSR}
B. {Lim}, S. {Son}, H. {Kim}, S. {Nah}, and K.~M. {Lee}.
\newblock Enhanced deep residual networks for single image super-resolution.
\newblock In {\em 2017 IEEE Conference on Computer Vision and Pattern
  Recognition Workshops (CVPRW)}, pages 1132--1140, 2017.

\bibitem{RFDN}
Jie Liu, Jie Tang, and Gangshan Wu.
\newblock Residual feature distillation network for lightweight image
  super-resolution.
\newblock In {\em European Conference on Computer Vision}, pages 41--55.
  Springer, 2020.

\bibitem{SCNet}
Jiang-Jiang Liu, Qibin Hou, Ming-Ming Cheng, Changhu Wang, and Jiashi Feng.
\newblock Improving convolutional networks with self-calibrated convolutions.
\newblock In {\em Proceedings of the IEEE/CVF Conference on Computer Vision and
  Pattern Recognition}, pages 10096--10105, 2020.

\bibitem{B100}
D. {Martin}, C. {Fowlkes}, D. {Tal}, and J. {Malik}.
\newblock A database of human segmented natural images and its application to
  evaluating segmentation algorithms and measuring ecological statistics.
\newblock In {\em Proceedings Eighth IEEE International Conference on Computer
  Vision. ICCV 2001}, volume~2, pages 416--423 vol.2, 2001.

\bibitem{manga109}
Yusuke Matsui, Kota Ito, Yuji Aramaki, Toshihiko Yamasaki, and Kiyoharu Aizawa.
\newblock Sketch-based manga retrieval using manga109 dataset.
\newblock 2015.

\bibitem{deblur}
J. {Pan}, W. {Ren}, Z. {Hu}, and M. {Yang}.
\newblock Learning to deblur images with exemplars.
\newblock {\em IEEE Transactions on Pattern Analysis and Machine Intelligence},
  41(6):1412--1425, 2019.

\bibitem{ESPCNN}
W. {Shi}, J. {Caballero}, F. {Huszár}, J. {Totz}, A.~P. {Aitken}, R. {Bishop},
  D. {Rueckert}, and Z. {Wang}.
\newblock Real-time single image and video super-resolution using an efficient
  sub-pixel convolutional neural network.
\newblock In {\em 2016 IEEE Conference on Computer Vision and Pattern
  Recognition (CVPR)}, pages 1874--1883, 2016.

\bibitem{DRRN}
Y. {Tai}, J. {Yang}, and X. {Liu}.
\newblock Image super-resolution via deep recursive residual network.
\newblock In {\em 2017 IEEE Conference on Computer Vision and Pattern
  Recognition (CVPR)}, pages 2790--2798, 2017.

\bibitem{MemNet}
Y. {Tai}, J. {Yang}, X. {Liu}, and C. {Xu}.
\newblock Memnet: A persistent memory network for image restoration.
\newblock In {\em 2017 IEEE International Conference on Computer Vision
  (ICCV)}, pages 4549--4557, 2017.

\bibitem{DIV2K}
R. {Timofte}, E. {Agustsson}, L.~V. {Gool}, M. {Yang}, L. {Zhang}, B. {Lim}, S.
  {Son}, H. {Kim}, S. {Nah}, and K.~M.~{Lee} et al.
\newblock Ntire 2017 challenge on single image super-resolution: Methods and
  results.
\newblock In {\em 2017 IEEE Conference on Computer Vision and Pattern
  Recognition Workshops (CVPRW)}, pages 1110--1121, 2017.

\bibitem{Flickr2K}
Radu Timofte, Eirikur Agustsson, Luc Van~Gool, Ming-Hsuan Yang, Lei Zhang, Bee
  Lim, et~al.
\newblock Ntire 2017 challenge on single image super-resolution: Methods and
  results.
\newblock In {\em The IEEE Conference on Computer Vision and Pattern
  Recognition (CVPR) Workshops}, July 2017.

\bibitem{Transrecog2}
Hugo Touvron, Matthieu Cord, Matthijs Douze, Francisco Massa, Alexandre
  Sablayrolles, and Herv{\'e} J{\'e}gou.
\newblock Training data-efficient image transformers \& distillation through
  attention.
\newblock In {\em International Conference on Machine Learning}, pages
  10347--10357. PMLR, 2021.

\bibitem{valanarasu2021transweather}
Jeya Maria~Jose Valanarasu, Rajeev Yasarla, and Vishal~M Patel.
\newblock Transweather: Transformer-based restoration of images degraded by
  adverse weather conditions.
\newblock {\em arXiv preprint arXiv:2111.14813}, 2021.

\bibitem{Seg1}
Wenhai Wang, Enze Xie, Xiang Li, Deng-Ping Fan, Kaitao Song, Ding Liang, Tong
  Lu, Ping Luo, and Ling Shao.
\newblock Pyramid vision transformer: A versatile backbone for dense prediction
  without convolutions.
\newblock In {\em Proceedings of the IEEE/CVF International Conference on
  Computer Vision}, pages 568--578, 2021.

\bibitem{NLRN}
Xiaolong Wang, Ross Girshick, Abhinav Gupta, and Kaiming He.
\newblock Non-local neural networks.
\newblock In {\em Proceedings of the IEEE Conference on Computer Vision and
  Pattern Recognition (CVPR)}, June 2018.

\bibitem{A2F}
Xuehui Wang, Qing Wang, Yuzhi Zhao, Junchi Yan, Lei Fan, and Long Chen.
\newblock Lightweight single-image super-resolution network with attentive
  auxiliary feature learning.
\newblock In {\em Proceedings of the Asian conference on computer vision},
  2020.

\bibitem{Seg2}
Enze Xie, Wenhai Wang, Zhiding Yu, Anima Anandkumar, Jose~M Alvarez, and Ping
  Luo.
\newblock Segformer: Simple and efficient design for semantic segmentation with
  transformers.
\newblock {\em Advances in Neural Information Processing Systems}, 34, 2021.

\bibitem{SRT}
Fuzhi Yang, Huan Yang, Jianlong Fu, Hongtao Lu, and Baining Guo.
\newblock Learning texture transformer network for image super-resolution.
\newblock In {\em Proceedings of the IEEE/CVF conference on computer vision and
  pattern recognition}, pages 5791--5800, 2020.

\bibitem{restormer}
Syed~Waqas Zamir, Aditya Arora, Salman Khan, Munawar Hayat, Fahad~Shahbaz Khan,
  and Ming-Hsuan Yang.
\newblock Restormer: Efficient transformer for high-resolution image
  restoration.
\newblock {\em arXiv preprint arXiv:2111.09881}, 2021.

\bibitem{Set14}
Roman Zeyde, Michael Elad, and Matan Protter.
\newblock On single image scale-up using sparse-representations.
\newblock 2010.

\bibitem{AIM2020}
Kai Zhang, Martin Danelljan, Yawei Li, Radu Timofte, Jie Liu, Jie Tang,
  Gangshan Wu, Yu Zhu, Xiangyu He, Wenjie Xu, et~al.
\newblock Aim 2020 challenge on efficient super-resolution: Methods and
  results.
\newblock In {\em European Conference on Computer Vision}, pages 5--40.
  Springer, 2020.

\bibitem{degrade1}
Kai Zhang, Wangmeng Zuo, Shuhang Gu, and Lei Zhang.
\newblock Learning deep cnn denoiser prior for image restoration.
\newblock In {\em Proceedings of the IEEE conference on computer vision and
  pattern recognition}, pages 3929--3938, 2017.

\bibitem{degrade2}
Kai Zhang, Wangmeng Zuo, and Lei Zhang.
\newblock Learning a single convolutional super-resolution network for multiple
  degradations.
\newblock In {\em Proceedings of the IEEE conference on computer vision and
  pattern recognition}, pages 3262--3271, 2018.

\bibitem{RCAN}
Yulun Zhang, Kunpeng Li, Kai Li, Lichen Wang, Bineng Zhong, and Yun Fu.
\newblock Image super-resolution using very deep residual channel attention
  networks.
\newblock 2018.

\bibitem{PAN}
Hengyuan Zhao, Xiangtao Kong, Jingwen He, Yu Qiao, and Chao Dong.
\newblock Efficient image super-resolution using pixel attention.
\newblock In {\em European Conference on Computer Vision}, pages 56--72.
  Springer, 2020.

\bibitem{PSNR}
{Zhou Wang}, A.~C. {Bovik}, H.~R. {Sheikh}, and E.~P. {Simoncelli}.
\newblock Image quality assessment: from error visibility to structural
  similarity.
\newblock {\em IEEE Transactions on Image Processing}, 13(4):600--612, 2004.

\bibitem{NAS}
Barret Zoph and Quoc~V Le.
\newblock Neural architecture search with reinforcement learning.
\newblock {\em arXiv preprint arXiv:1611.01578}, 2016.

\end{thebibliography}
}

\end{document}